\documentclass[conference]{IEEEtran}

  	\usepackage[pdftex]{graphicx}
  	\graphicspath{{../pdf/}{../jpeg/}}
	\DeclareGraphicsExtensions{.pdf,.jpeg,.png}

	\usepackage[cmex10]{amsmath}
	\usepackage{amsfonts}
	\usepackage{array}
	\usepackage{url}
    \usepackage{xspace}
    \usepackage{hyperref}
    
    \IEEEoverridecommandlockouts

\begin{document}

\title{\LARGE {Development of Human Motion Prediction Strategy  using  Inception Residual Block}}

\author{\IEEEauthorblockN{Shekhar Gupta, Gaurav Kumar Yadav, G. C. Nandi}
\text{Center of Intelligent Robotics} \\
\text{Indian Institute of Information Technology Allahabad}\\
\text{Prayagraj, UP-211015, INDIA} \\
\text{Email: guptashekhar54@gmail.com}
}
\maketitle

\begin{abstract}
Human Motion Prediction is a crucial task in computer vision and robotics. It has versatile application potentials such as in the area of human-robot interactions, human action tracking for airport security systems, autonomous car navigation, computer gaming to name a few. However, predicting human motion based on past actions is an extremely challenging task due to the difficulties in detecting spatial and temporal features correctly. To detect temporal features in human poses, we propose an Inception Residual Block(IRB), due to its inherent capability of processing multiple kernels to capture salient features. Here, we propose to use multiple 1-D Convolution Neural Network (CNN) with different kernel sizes and input sequence lengths and concatenate them to get proper embedding. As kernels strides over different receptive fields, they detect smaller and bigger salient features at multiple temporal scales. Our main contribution is to propose a residual connection between input and the output of the inception block to have a continuity between the previously observed pose and the next predicted pose. With this proposed architecture, it learns prior knowledge much better about human poses and we achieve much higher prediction accuracy as detailed in the paper. Subsequently, we further propose to feed the output of the inception residual block as an input to the Graph Convolution Neural Network (GCN) due to its better spatial feature learning capability. We perform a parametric analysis for better designing of our model and subsequently, we evaluate our approach on the Human 3.6M dataset and compare our short-term as well as long-term predictions with the state of the art papers, where our model outperforms most of the pose results, the detailed reasons of which have been elaborated in the paper. 
\end{abstract}

\IEEEoverridecommandlockouts


\IEEEpeerreviewmaketitle

\begin{figure*} 
\centering
\includegraphics[width=7.0in]{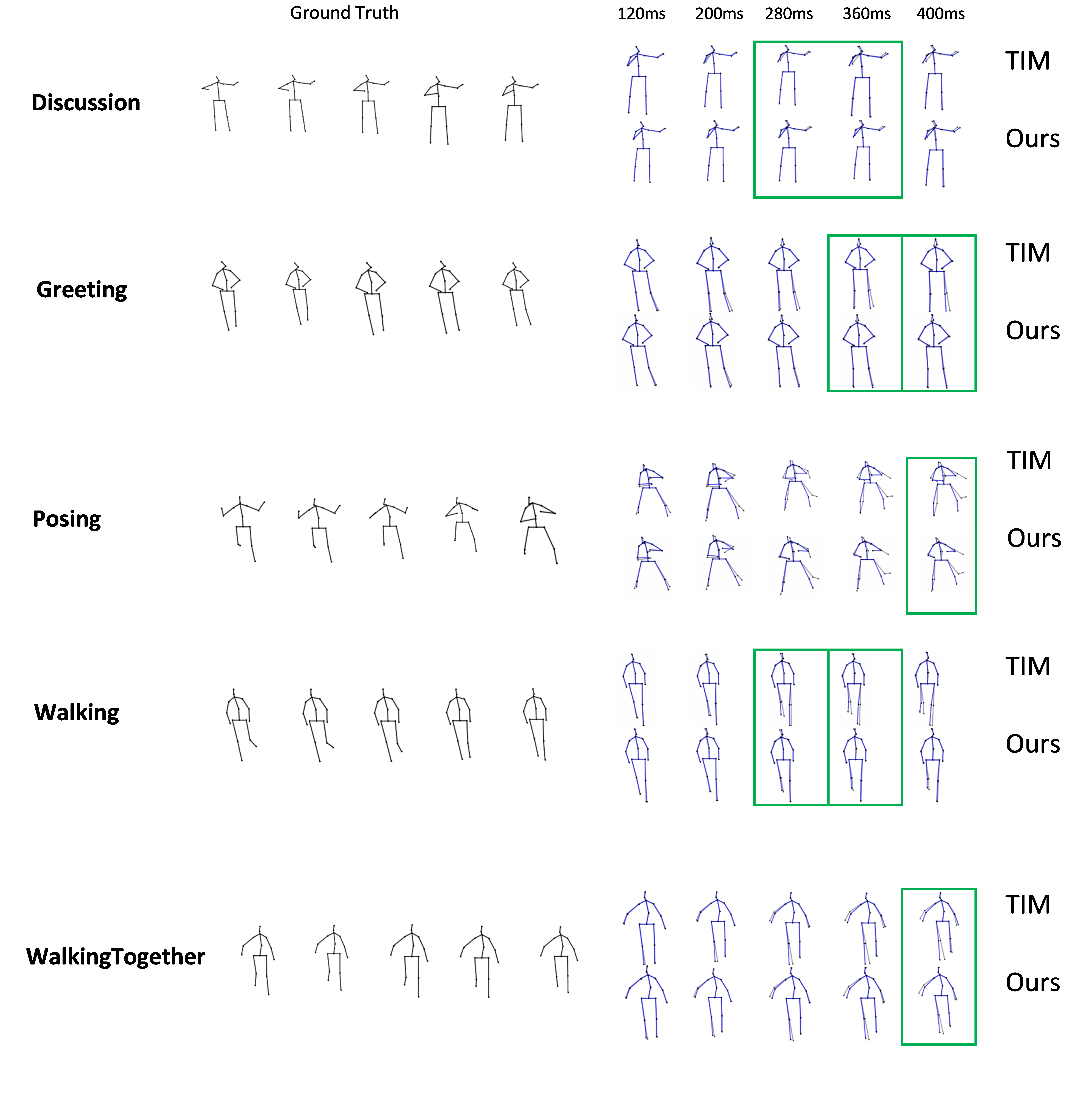}
\caption{Pose comparison between TIM model and ours: For a activity On the left side we have shown ground truth(in black color). On the right side we have shown prediction. TIM model prediction shown in 1st row and ours model is shown in 2nd row(written at end). We have superimposed ground truth with both methods where blue pose is predicted and black pose is ground truth. Some of our best prediction comparison with TIM model are highlighted. 
}
\label{fig:1}
\end{figure*}

\section{Introduction}
Human motion prediction is a crucial task where the machine needs to predict human pose. Recognizing human pose tells human behaviour to machine. It involves human tracking \cite{gong2011multi}, \cite{gupta20143d} \cite{10.1007/11789239_18}, motion generation \cite{kovar2008motion}, human action recognition and prediction\cite{kong2018human},  online gaming \cite{lau2008motion}. Generally it is used where machines need to interact with humans and useful in computer vision and robotics. In human-robot interaction, \cite{koppula2013anticipating} if a human is walking and a robot have to approach a human the robot must predict human motion. In Autonomous car driving \cite{paden2016survey}, car must have to predict pedestrian motion \cite{habibi2018context} to prevent accident. 

Our task in this paper is to predict human motion and achieving higher accuracy. We have given past human pose and need to forecast next human pose. To achieve higher accuracy key point is to detect temporal features. Authors in \cite{fragkiadaki2015recurrent}, \cite{martinez2017human} \cite{yadav2020development} use RNNs to detect temporal features. There are two problems with the RNN based models. First problem \cite {fragkiadaki2015recurrent} \cite{martinez2017human} is, in RNN, errors adds up to every step of sequences. It leads to illogical prediction at testing time. Second, as it is found in \cite{fragkiadaki2015recurrent} and  \cite{martinez2017human}, there is an inconsistency between the last observed and first forecasted frames. This inconsistency exists because of the frame by frame regression that does not encourage global smoothness. 

Mao et. al \cite{mao2019learning} considers this task as feed forward network. They have used Discrete Cosine Transform(DCT) to encode temporal features and graph convolution network \cite{velivckovic2017graph} as feed-forward network for prediction. Tim et. al \cite{10.1007/978-3-030-69532-3_39} follows \cite{mao2019learning} research and instead of DCT uses inception module \cite{szegedy2015going}. In line with \cite{10.1007/978-3-030-69532-3_39}, for our specific problem for predicting human poses based on past actions, we hypothesise that using residual block with inception module can provide better continuity between previously observed pose and the next predicted pose. In the inception module, we are proposing to use 1D CNN since 1D CNN plays a crucial role to detect temporal features.

CNN \cite{lecun1998gradient} considers three architectural idea: local receptive field, shared weights and spatial sub-sampling. Receptive field slides over data by the predefined value that is called stride.  While sliding they overlap with each other and cover the whole series of data and detects features of input data. CNN shares weight and bias for a layer which makes it computational cost-efficient than artificial neural network. Weights and bias learn by training. By doing one convolution operation one feature is detected. When kernel strides in 1-dimensional time series data it detects temporal features of input data.

In our work to detect temporal features, we applied the inception module and the residual block. In the inception module \cite {szegedy2015going}  we apply different size of kernels to get different receptive fields so that salient features can be detected. By considering multiple sizes of kernels we have two advantages. First, we don't have to choose perfect kernel size for input and makes worry free. Second, model gets a better look at input data. It detects smaller to bigger key features. Adding residual block makes it easier to learn residual mapping than original mapping \cite{he2016deep}. It adds continuity between previously observed pose and temporal encoding of pose. Using the Inception residual module we encode human poses to temporal encoding. 

Now the question is how to detect spatial features of a human pose. In our work spatial features means learning dependencies between human joints of pose. It is important that our approach should not depend on fixed convolution filter size which is similar to detecting temporal features. We are using GCN \cite{kipf2016semi} to learn graph connectivity and detecting spatial features. GCNs are powerful feed forward neural network that learns human pose joint dependencies \cite{mao2019learning}. As an input they take temporal encoding and predict human motion. 

The major contributions of the paper: 
\begin{itemize}
    \item We have proposed a residual connection between input and inception block.
    \item We have done parametric analyses of number of temporal features.
    \item We perform end-to-end learning with Inception Residual Block and GCN. 
\end{itemize}

This paper has divided into five parts. In the First Part, we have discussed about human motion prediction and talked about its application, problem analysis as well as gave a little overview of 1D CNN. In the second part, we have discussed related work. In the third part, we have shared our methodology where we formulate a problem statement, discuss preliminaries, and explained our architecture and training details. In the Fourth part, we have shown our implementation details, discussed parametric analysis and compared our result with baseline methods . In the final part, we have written our conclusion and future possibilities. 

\section{Related Work}
\subsection{RNN based models forecasts human motion prediction}
RNNs are default choice to detect time-dependent features. \cite{kiros2015skip} \cite{sutskever2011generating} considers human motion prediction as sequence to sequence prediction. \cite{fragkiadaki2015recurrent} have introduced 2 approaches: LSTM-3LR and Encoder-Recurrent-Decoder(ERD). In both architectures, they have used a layer of LSTM cells. But in ERD architecture they have added a non-linear encoder for data pre-processing. To prevent error accumulation they added noise in the input. But adding noise made long-term prediction tough and inconsistent. Sang et. al \cite{sang2020human} also proposed 2 models: At-seq2seq and seq2seq. In At-seq2seq model with attention mechanism They have used GRU cell in encoder and decoder. In seq2seq they did not use attention mechanism and got better result with attention mechanism. Jain et. al \cite{jain2016structural} have suggested S-RNN(structural). Using S-RNN they have made spatio-temporal graph and detected spatial and temporal features of human pose. But they have designed graphs manually so they were not flexible and suffers from lack of long-term dependencies. Martinez et. al \cite{martinez2017human} have proposed a simple baseline method. They have also used sequence to sequence architecture and added residual connection between the input and output of RNN cells. Surprisingly their approach outperforms most of the previous solutions. Gaurav et. al \cite{yadav2020development} have applied adaptive sampling based cost function instead of sampling based loss function and got a better result for many poses than \cite{martinez2017human}.

\subsection{Feed-forward approaches forecasts human motion}
To encounter discontinuity in RNNs, fully-connected network \cite{butepage2017deep} and convolutions \cite{li2018convolutional} approaches studied. Butepage et. al \cite{butepage2017deep} have modeled human poses as input to fully connected layers and applied different strategies to detect temporal features. To detect spatial features they have used kinematic trees. They have built tree on specific dataset which made it inflexible. So kinematic tree have not synced with different body parts such as limbs. Li et. al \cite{li2018convolutional} have introduced convolutional sequence to sequence. Spatial and temporal features have captured by convolution kernels. They have fixed kernel size so they were also inflexible and couldn't detect features properly.    

Kjellstr{\'o}m et. al \cite{butepage2018anticipating} have suggested Conditional Variational Autoencoder(CVAE), a probabilistic approach  to predict and generate human motion. Based on CVAE Kragic et. al \cite{butepage2019predicting} have proposed semi-supervised recurrent neural network(SVRNN) to detect or classify and predict human pose. Aliakbarian et. al \cite{aliakbarian2020stochastic} have used recurrent-encoder-decoder network with CVAE for prediction. They have used future pose autoencoder with CVAE and placed in between encoder and decoder.    

\subsection{Graph Convolution Network}
GCN captures information passing through node in graph. It involves mainly node classification \cite{kipf2016semi} \cite{monti2017geometric} \cite{hamilton2017inductive}, image classification \cite{monti2017geometric} \cite{defferrard2016convolutional}, machine translation problem \cite{marcheggiani2018exploiting}, recommendation system \cite{monti2017geometric1} \cite{ying2018graph}. GCN do convolution operation on graph data-structure in two ways defined in \cite{kipf2016semi} and \cite{velivckovic2017graph}. Kipf et. al \cite{kipf2016semi} have used layer-wise propagation rule for nodes which was inspired from first order approximation of spectral convolutions on graphs. They have applied convolution operation based on the graph structure. It is limited to the characteristics of the graph. While Veli{\v{c}}kovi{\'c} et. al \cite{velivckovic2017graph} have learned node connectivity using neighbourhood of node. So it is more flexible to GCN. So we are using GCN architecture motivated by \cite{velivckovic2017graph} to learn node embedding adaptively.

\begin{figure*} 
\centering
\includegraphics[width=7.0in]{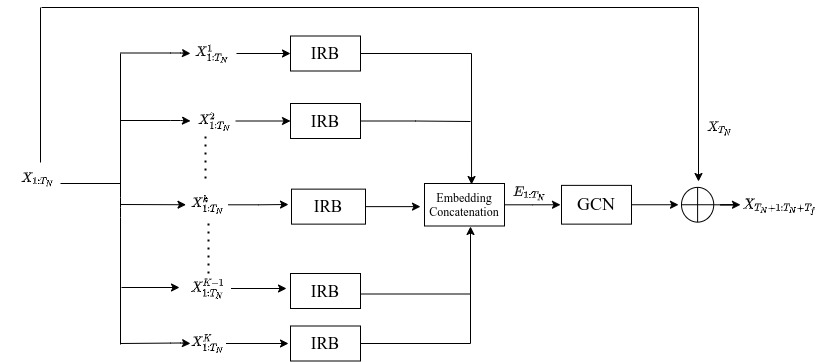}
\caption{The full architecture of our proposed method: We break human poses $X_{1:T_{N}}$ into K human joint ${{X^1_{1:N}} ... {X^{K}_{1:N}}}$ trajectory. Each ${X^k_{1:T_{N}}}$ pose fed to IRB(Fig. 3) and generate ${E^k_{1:T_{N}}}$ embedding. All embedding are concatenated and made node features ${E_{1:T_{N}}}$ of the graph which fed to GCN and added with ${X_N}$ most recent pose.}
\label{fig:2}
\end{figure*}
\subsection{Inception Block}
Inception Module has been introduced first time in the ILSVRC14 image classification competition by szegedy et. al\cite{szegedy2015going}. They use in GooleLeNet and their result proved that using inception block one can achieve higher accuracy. It makes the network wider rather than deeper which helps to reduce computation cost. Ioffe et. al \cite{szegedy2016rethinking} propose 4 design principles to build CNN architecture. Based on these principles they introduce inception v2 and v3 architecture and got a better result. Sergey et. al \cite{szegedy2017inception} suggested inception-v4. It was a little bit complex than the previous ones but gave a good result. We are using simple inception architecture \cite{szegedy2015going} without max-pooling layer.

\subsection{Baseline Methods}
We are comparing our results in MPJPE(Mean Per Joint Position Error) Error \cite{ionescu2013human3} because all of the baseline methods which we are considering is also using MPJPE metric. To compare our result we are using following baselines: 2 famous sequence to sequence based methods. First, Martinez et. al \cite{martinez2017human} used well-known RNN method and Li et. al \cite{li2018convolutional} used Convolution to encode and decode data. Mao et. al \cite{mao2019learning} and Tim et. al \cite{10.1007/978-3-030-69532-3_39} both encode human poses. Mao et. al \cite{mao2019learning} used DCT + GCN and Tim et. al \cite{10.1007/978-3-030-69532-3_39} used TIM + GCN. We didn't compare our result with \cite{li2020dynamic} \cite{yadav2020development} papers because they used different metric mean angle error.

\section{Methodology}
\subsection{Problem Statement}
We are using human pose 3D joint positions. For each joint, we have its x,y,z coordinate. We have given time-series ${T_{N}}$ 3D-joint coordinates as shown in equation 1. 

\begin{equation}
 {X_{1:T_{N}}}  = [{X_{1}}, {X_{2}}, ... {X_{T_{N}}}]   
\end{equation}

Where $X_{1:T_{N}} \in \mathbb{R} ^ {T_{N} * K * D}$, $T_{N}$ is the number of input time steps, K is number of joints in human pose and ${D}$ = 3 is feature dimension x,y,z. Our task is to predict ${X_{T_{N}+1 : {T_{N}+{T_{f}}}}}$, continuous human poses, where ${T_{f}}$ number of future time steps.

\begin{figure} 
\centering
\includegraphics[width=3in]{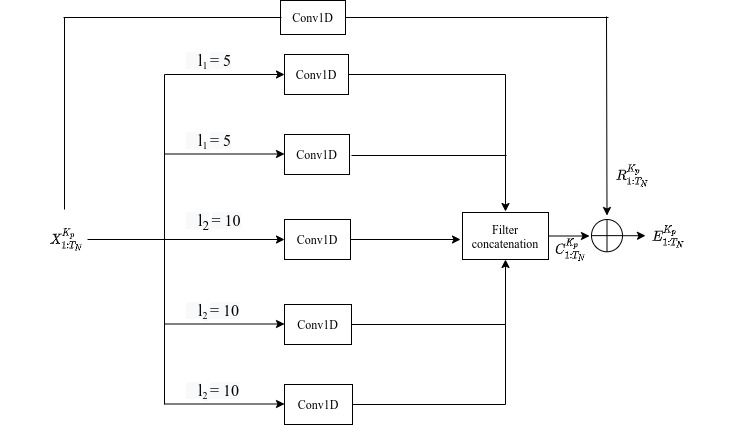}
\caption{Full architecture of Inception Residual Block: For ${k}$th joint, feature dimension $p$ $\in$ $\{x,y,z\}$, ${X_{1:T_{N}}^{k_{p}}}$ is fed to 5 Conv1d filters with different input length and their output is concatenated and generates ${C_{1:T_{N}}^{k_{p}}}$. To match ${X_{1:T_{N}}^{k_{p}}}$ same shape whole feature dimension trajectory is passed to kernel size 1 Conv1d, added with ${C_{1:T_{N}}^{k_{p}}}$ and produces feature dimension embedding ${E_{1:T_{N}}^{k_{p}}}$}
\label{fig:3}
\end{figure}

\subsection{Proposed Approach}
We are making a graph represented by an adjacency matrix where the nodes are represented by human joints of pose. To calculate node features we propose an Inception Residual Block(IRB). These node features are fed to GCN. GCN learns node features of graph and produce the future expected poses ${X_{T_{N}+1 : {T_{N}+{T_{f}}}}}$. Details of network architecture we have discussed in the following sections.

\subsection{Inception Module}
When we increase the depth of CNN, the model becomes more susceptible to overfitting \cite{szegedy2015going}. As parameters increases model becomes harder to train. Choosing the right kernel size is also a problem because in some training examples salient features are big while in some are smaller. To encounter all of these problems at once Szegedy et. al \cite{szegedy2015going} applies multiple kernels parallel in the model. In the inception module \cite{szegedy2015going}, kernel size used 1, 3, 5 and max-pooling layer. Using multiple kernels models can capture both bigger and smaller salient features. By concatenation, the model becomes wider so computationally inexpensive and overcomes overfitting. 

\subsection{Temporal Encoding using Inception Residual Block}
We have shown Inception Residual Block(IRB) module in figure 3. The main purpose of this block is to detect temporal features of feature dimension trajectory (x, y, z) of the human joint. It takes feature dimension trajectory ${X_{1:T_{N}}^{k_{p}}}$ and generates embedding ${E_{1:T_{N}}^{k_{p}}}$. 

IRB takes two different size ${l_{1}}$ = 5 and ${l_{2}}$ = 10 input length feature dimension trajectory. In inception block, For ${l_{1}}$ length, we apply 2 1D convolution operation and for ${l_{2}}$ length 3 convolution operation. On each input, different size kernels are applied. For more details of kernel size and number of kernels refer to Table 1. As Tim et. al \cite{10.1007/978-3-030-69532-3_39} suggested for shorter input length shorter kernel size and larger input length larger kernel size applied. Small kernel detects small salient and large kernel detects big salient features. All 1D CNN output are concatenated into one embedding ${C_{1:T_{N}}^{k_{p}}}$. 

To match same shape ${X_{1:T_{N}}^{k_{p}}}$ with ${C_{1:T_{N}}^{k_{p}}}$ we apply 1D convolution with kernel size 1 and getting ${R_{1:T_{N}}^{k_{p}}}$. We have added residual connection between ${R_{1:T_{N}}^{k_{p}}}$ and  ${C_{1:T_{N}}^{k_{p}}}$. In other words, between input and the output of the inception block. There are 2 main advantages. First, to remove discontinuity between the previously observed pose and the next predicted pose in addition it helps to learn prior knowledge much better about human poses. Second,  during backpropagation algorithm gradient can flow through the model directly which prevents gradients from vanishing or exploding. So, residual connection \cite{he2016deep} improves performance. Finally residual connection generates embedding ${E_{1:T_{N}}^{k_{p}}}$.

\begin{figure*} 
\centering
\includegraphics[width=1.0\textwidth]{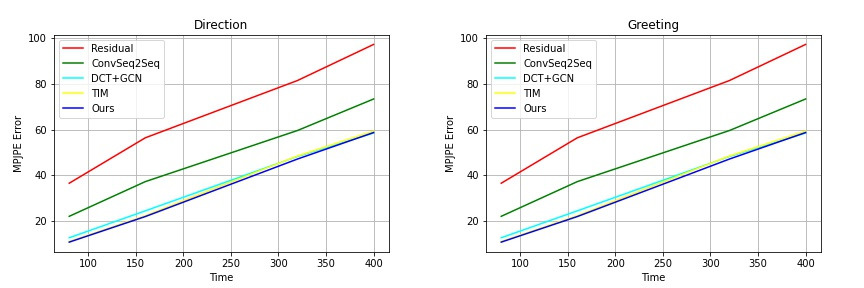}

\caption{Comparison of MPJPE error for Direction and Greeting pose with "Our" approach and Baseline methods}
\label{fig:4}
\end{figure*}

\subsection {Graph Convolution Network}
To predict human motion we are using the same GCN as suggested by \cite{mao2019learning}. A human 3D skeleton can interpret as a graph with ${K}$ fully connected nodes, where ${K}$ is several body joints. The graph is denoted by ${A \in}$ ${\mathbb{R} ^{K * K}}$ adjacency matrix where the nodes are human joints. Graph convolution layer takes hidden feature matrix ${H \in}$ ${\mathbb{R} ^{K * F}}$, where ${F}$ is features of the previous layer. In the first layer hidden matrix is output of Inception residual block output. In Graph Convolution layer each node is accumulation of node neighbour features. We use multiple stacked graph convolution layers. Each layer performs a given operation shown in equation 2.  

\begin{equation}
    H^{(p+1)} = \sigma(A^{(p)} H^{(p)} W^{(p)})
\end{equation}

Here ${\sigma(.)}$ is the activation function and ${A^{(p)}}$ is  adjacency matrix for layer $p$. ${H^{(p)}}$ is  hidden feature matrix for layer $p$. $W^{(p)}$ is the weight matrix of layer ${p}$ where ${W^{(p)} \in}$ ${\mathbb{R} ^{F * F}}$. ${H^{(p + 1)}}$ is output hidden feature matrix for layer $(p)$.

Both ${A}$ and ${W}$ are learnable matrices. Learnable matrix leads better result it is proven by mao et. al \cite{mao2019learning}. Both ${A}$ and ${W}$ are trained using a standard backpropagation algorithm.  

GCN takes Inception residual block generated embedding ${E_{1:T_{N}}}$. GCN learns node features of graph. As it learns node features it detects spatial features of graph. Output of GCN will be added with residual block as the most recent human pose ${X_{N}}$ is shown in Fig 2. It gives our predicted output ${X_{T_{N} + 1 : {T_{N} + {T_{f}}}}}$. 

\begin{table*}
    \centering
    \scalebox{1.2}{%
    \begin{tabular}{|c |c | c | c | c|}
    \hline 
    Row No. & Sequence input length & Number of kernels & Kernel size & Number of features \\
    \hline
    1 & 5 & 17 & 2 & 17*4 \\
    2 & 5 & 16 & 3 & 16*3 \\ 
    3 & 10 & 14 & 3 & 14*8 \\
    4 & 10 & 13 & 5 & 13*6 \\
    5 & 10 & 11 & 7 & 11*4 \\
    6 & 10 & - & - & 10\\
    \hline
    7 & - & - & - & 360 \\
    \hline
    \end{tabular}}
 \caption{Details of Inception Block in IRB}
   \label{tab:1}
\end{table*}

\begin{table*}
    \centering
    \scalebox{1.25}{%
    \begin{tabular}{ |c | c | c | c | c| }
    \hline 
    Motion & Walking & Eating & Smoking & Discussion \\ 
    \hline
    milliseconds & 80 160  320  400 & 80  160  320  400 & 80  160 320  400 & 80  160  320  400 \\
    \hline
    Residual\cite{martinez2017human} & 23.8 40.4 62.9 70.9 & 17.6 34.7 71.9 87.7 & 19.7 36.6 61.8 73.9 & 31.7 61.3 96.0 103.5 \\
    ConSeq2Seq \cite{jain2016structural} & 17.1 31.2 53.8 61.5 & 13.7 25.9 52.5 63.3 & 11.1 21.0 33.4 38.3 & 18.9 39.3 67.7 75.7 \\
    DCT + GCN \cite{mao2019learning} & \textbf{8.9} 15.7 29.2 33.4 & 8.8 18.9 39.4 47.2 & 7.8 14.9 25.3 28.7 & 9.8 22.1 39.6 44.1\\
    
    TIM\cite{10.1007/978-3-030-69532-3_39} & 9.3 15.9 30.1 34.1 & 8.4 18.5 \textbf{38.1 46.6} & \textbf{6.9} 13.8 24.6 29.1 & 8.8 21.3 40.2 45.5\\
    \hline
    Ours & \textbf{8.9 15.2} \textbf{29.1 32.7} & \textbf{8.1 18.3} 39.1 47.0 & \textbf{6.9 13.6 23.9 28.2} & \textbf{8.4 19.7 36.1 40.7} \\
    \hline
    \end{tabular}}
 \caption{Detail results of short-term prediction, measure in MPJPE error of different poses, e.g., Walking, Eating, Smoking, and Discussing.}
   \label{tab:2}
\end{table*}

\begin{table*}
    \centering
    \scalebox{1.25}{%
    \begin{tabular}{ |c | c | c | c | c| }
    \hline 
    Motion & Directions & Greeting & Phoning & Posing  \\ 
    \hline
    milliseconds & 80 160  320  400 & 80  160  320  400 & 80  160 320  400 & 80  160  320  400 \\
    \hline
    Residual\cite{martinez2017human} & 36.5 56.4 81.5 97.3 & 37.9 74.1 139.0 158.8 & 25.6 44.4 74.0 84.2 & 27.9 54.7 131.3 160.8 \\
    ConvSeq2Seq \cite{jain2016structural} & 22.0 37.2 59.6 73.4 & 24.5 46.2 90.0 103.1 & 17.2 29.7 53.4 61.3 & 16.1 35.6 86.2 105.6 \\
    DCT + GCN \cite{mao2019learning} & 12.6 24.4 48.2 \textbf{58.4} & 14.5 30.5 74.2 89.0 & 11.5 20.2 37.9 \textbf{43.2} & 9.4 23.9 66.2 82.9\\
    
    TIM\cite{10.1007/978-3-030-69532-3_39} & 11.0 22.3 48.4 59.3 & \textbf{13.7} 29.1 72.6 \textbf{88.9} & 11.5 19.8 38.5 44.4 & 7.5 22.3 64.8 80.8\\
    \hline
    Ours & \textbf{10.7 21.9 47.1} 58.7 & \textbf{13.7 28.7 71.6} 89.7 & \textbf{11.2 19.4 37.5} 43.9 & \textbf{7.2 21.1 57.9 72.1}\\
    
    \hline
    \end{tabular}}
    \caption{Detail results of short-term prediction, measure in MPJPE error of different poses, e.g., Direction, Greeting, Phoning, and Posing.}
    \label{tab:3} 
\end{table*}

\begin{table*}
    \centering
    \scalebox{1.25}{%
    \begin{tabular}{ |c | c | c | c | c| }
    \hline 
    Motion & Purchases & Sitting & Sitting Down & Taking Photo \\ 
    \hline
    milliseconds & 80 160  320  400 & 80  160  320  400 & 80  160 320  400 & 80  160  320  400 \\
    \hline
   Residual\cite{martinez2017human} & 40.8 71.8 104.2 109.8 & 34.5 69.9 126.3 141.6 & 28.6 55.3 101.6 118.9 & 23.6 47.4 94.0 112.7\\
    ConvSeq2Seq \cite{jain2016structural} & 29.4 54.9 82.2 93.0 & 19.8 42.4 77.0 88.4 & 17.1 34.9 66.3 77.7 & 14.0 27.2 53.8 66.2\\
    DCT + GCN \cite{mao2019learning} & 19.6 38.5 64.4 \textbf{72.2} & 10.7 24.6 50.6 62.0 & 11.4 27.6 56.4 67.6 & 6.8 15.2 \textbf{38.2} 49.6\\
    TIM\cite{10.1007/978-3-030-69532-3_39} & 19.0 39.2 65.9 74.6 & \textbf{9.3 22.3 45.3 56.0} & 11.3 28.0 54.8 64.8 & \textbf{6.4} 15.6 41.4 53.5\\
    \hline
    Ours & \textbf{18.4 37.2 64.3} 74.5 & \textbf{9.3} 22.8 47.0 58.5 & \textbf{10.5 25.5} \textbf{49.6 60.0} & \textbf{6.4 15.2} 39.9 \textbf{51.6}\\
    \hline
    \end{tabular}}
        \caption{Detail results of short-term prediction, measure in MPJPE error of different poses, e.g., Purchases, Sitting, Sitting down, and Taking a photo.}
    \label{tab:4}
\end{table*}

\begin{table*}
    \centering
    \scalebox{1.25}{%
    \begin{tabular}{| c | c | c | c | c |}
    \hline 
    Motion & Waiting & Walking Dog & Walking Together & Average \\ 
    \hline
    milliseconds & 80 160  320  400 & 80  160  320  400 & 80  160 320  400 & 80  160  320  400 \\
    \hline
    Residual\cite{martinez2017human} & 29.5 60.5 119.9 140.6 & 60.5 101.9 160.8 188.3 & 23.5 45.0 71.3 82.8 & 30.8 57.0 99.8 115.5\\
    ConvSeq2Seq \cite{jain2016structural} & 17.9 36.5 74.9 90.7 & 40.6 74.7 116.6 138.7 & 15.0 29.9 54.3 65.8 & 19.6 37.8 68.1 80.2 \\
    DCT + GCN \cite{mao2019learning} & 9.5 22.0 57.5 73.9 & 32.2 58.0 102.2 122.7 & \textbf{8.9 18.4} 35.3 44.3 & 12.1 25.0 51.0 61.3\\
    TIM\cite{10.1007/978-3-030-69532-3_39} & 9.2 \textbf{21.7 55.9 72.1} & 29.3 \textbf{56.4} 99.6 119.4 & \textbf{8.9} 18.6 35.5 44.3 & 11.4 24.3 50.4 60.9\\
    \hline
    Ours & \textbf{9.0 21.7} 56.8 72.5 & \textbf{29.2} 58.3 \textbf{98.3 114.9} & 9.1 18.7 \textbf{34.2 43.0} & \textbf{11.1 23.8 48.8 59.2}\\
    \hline
    \end{tabular}}
      \caption{Detail results of short-term prediction, measure in MPJPE error of different poses, e.g., Waiting, Walking with the dog, Walking together, and  Average.}
    \label{tab:5}
\end{table*}

\begin{table*}
    \centering
    \scalebox{1.30}{%
    \begin{tabular}{ |c | c | c | c | c | c | }
    \hline 
    Motion & Walking & Eating & Smoking & Discussion & Average \\ 
    \hline
    milliseconds & 560 1000 & 560 1000 & 560 1000 & 560 1000 & 560 1000 \\
    \hline
    Residual\cite{martinez2017human} & 73.8 86.7 & 101.3 119.7 & 85.0 118.5 & 120.7 147.6 & 95.2 118.1 \\
    ConvSeq2Seq \cite{jain2016structural} & 59.2 71.3 & 66.5 85.4 & 42.0 67.9 & 84.1 116.9 & 62.9 85.4\\
    DCT + GCN \cite{mao2019learning} & 42.3 51.3 & 56.5 68.6 & 32.3 60.5 & 70.5 103.5 & 50.4 71.0 \\
    TIM\cite{10.1007/978-3-030-69532-3_39} & 39.6 46.9 & 56.9 68.6 & 33.5 61.7 & \textbf{68.5} 97.0 & 50.4 71.0 \\
    \hline
    Ours & \textbf{35.6 44.3} & \textbf{55.4 68.2}  & \textbf{31.1 56.0} & 68.8 \textbf{75.5} & \textbf{47.7 61.0}\\
    \hline
    \end{tabular}}
    \caption{For long term prediction, MPJPE ERROR OF DIFFERENT POSES of four actions and their average for H3.6M dataset.}
    \label{tab:6}
\end{table*}

\section{Experiments and Result analyses}

\subsection{Human 3.6M Dataset}
For human motion prediction, we are using the Human 3.6M dataset\cite{ionescu2013human3}. In this dataset there are 7 subjects are doing 15 actions walking, eating, smoking, walking together, etc. A human pose is represented by 32 joints. We use same data pre-processing \cite{martinez2017human} \cite{mao2019learning} and removed global rotations and translations. Similar to previous work \cite{mao2019learning}
\cite{li2018convolutional} \cite{martinez2017human} for training we used 5 subjects. 1 subject is used for checking the validation and one for testing. We are considering 25 frames per second. So, for 80ms we are considering one frame. For testing we use subject number 5 and similar to previous work same sequences are considered for testing.

\subsection{Implementation Details and Model Configuration}
We train our model to 50 epochs. For short-term prediction used 10 input and output frames. We followed the same work as \cite{mao2019learning} and set the learning rate to 0.0005 and it is decaying by 0.96 with every 2 epochs. For training our batch size is set to 16. For GCN we use 12 stacked convolution layers. For implementation we use PyTorch \cite{paszke2017automatic} and use ADAM optimizer \cite{kingma2014adam} and Tanh activation function. We run our code on NVIDIA Tesla V-100(16GB) GP-GPU(General purpose GPU). We conduct multiple experiments and choose the best result.

\subsection {Training Details}
For $K$ joints $K * 3$ Inception Residual Block(IRB) is used. In each IRB block, multiple different size 1D kernels are used. Smaller kernel like 2, 3 detect smaller features. For eg. in smoking pose only one hand is moving while other body part is still. Lower kernel size can easily detect smoking feature.  Larger kernel size like 5, 7 considers large receptive field. In walking pose whole body is moving both hands and legs. To detect continuously moving legs and hands larger kernel size are required.

We have used 1D convolution as we are feeding feature dimension trajectory at multiple temporal scales which is taking multiple length input ${l_{1}}$ = 5 and ${l_{2}}$ = 10. In table 1, we have shown details of IRB. In row 1, we take input sequence length 5 and apply kernel size 2, stride 1, zero-padding with 17 Conv1D kernels. We are concatenating Conv1D output so we are getting 17 * (5 - 2 + 1) features and row 7 concatenate input sequence length 10. After performing all concatenation we get temporal embedding 360.   
This 360 size embedding is added with input as a residual connection. Every IRB block embedding ${E_{1:T_{N}}^{k_{p}}}$ concatenated where $ k  \in K $ and $ p \in {x,y,z} $ (p = 3) \& ${E_{1:T_{N}}}$ fed to GCN. IRB and GCN are training for an end to end learning. 

For the loss function, we are using Mean Per Joint Position Error(MPJPE) \cite{ionescu2013human3} instead of mean square error for calculating the error between predicted pose and actual pose.
We also use this error for comparing our results with other states of the art results.
MPJPE is calculated by the following equation
\begin{equation}
    Error = \displaystyle\frac{1}{K(T_{f} +T_{N})} \sum_{n=1}^{T_{N} + T_{f}} \sum_{j=1}^{K*3}   ||\hat{p}_{j,n} - p_{j,n}||^2
\end{equation}
Where ${\hat{p}_{j,n}}$ ${\in}$ ${{R}^3}$ is predicted ${jth}$ joint position at ${nth}$ time frame and ${p_{j,n}}$ is corresponding ground-truth and K is number of joints in human pose(multiplied by 3 because x,y,z feature dimension).  

\subsection{Optimization of features with parametric analysis}

We train inception module with different number of kernels. As we are varying kernels we have considered temporal embedding features from 223 to 460. But on 460 features model becomes more prone to overfitting. In Fig 5. for 460 features and 420 features training error is very less while validation error is going very high that indicates overfitting. Same behaviour can be seen from table 7. For walkingdog3d400 prediction error is very high. 

For 223 and 300 features model was not overfit but can train more with features. Their validation error and training error are in same degree. This behaviour shown in Fig 5 and table 7. We have trained our model with 360 features which we found best. We have chosen 360 features because on some poses because we are getting better result for eg. in table 7 walking3d400, walkingdog3d400, walkingtogether3d400.

Based on numbers of features we have modified residual connection Conv1D kernels. We have changed only numbers of kernels while other parameters kept constant. In table 7 we have taken average over last 5 epochs.
\begin{figure*} 
\centering
\includegraphics[width=7.0in]{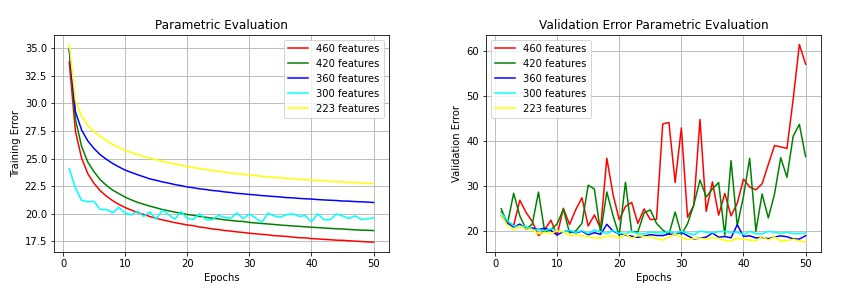}
\caption{Analysis of Training and Validation error based on number of temporal features}
\label{fig:5}
\end{figure*}

\begin{table*}
    \centering
    \scalebox{1.2}{%
    \begin{tabular}{|c |c | c | c|c|c|}
    \hline 
    No. of Features & Avg Training Loss & Avg Validation loss & walking3d400 & walkingdog3d400 & walkingtogether3d400 \\
    \hline
    223 & 22.8 & 18.0 & 35.4 & 118.8 & 45.0 \\
    300 & 19.4 & 19.6 & 35.3 & 114.8 & 45.2 \\
    \textbf{360} & \textbf{21.1} & \textbf{18.6} & \textbf{32.7} & \textbf{114.8} & \textbf{43.0} \\
    420 & 18.5 & 38.4 & 33.5 & 218.4 & 41.6 \\ 
    460 & 17.5 & 51.6 & 34.7 & 515.2 & 42.0 \\
    \hline
    \end{tabular}}
 \caption{Parametric Analysis of Number of Temporal features}
   \label{tab:7}
\end{table*}

\subsection{Result Analysis}
We have shown our results in table no. 2,3,4,5 for all poses.  We report our results over short-term prediction for 80, 160, 320, 400ms and use the same input length frame 10 and output length frame 10 to capture short-term prediction. We have taken average our result over the last 5 epochs. For most of the actions, we get a better result than baseline methods. In Fig. 4 comparison of MPJPE error with the short-term prediction time frame is further elaborated.  We conduct the parametric analysis and find that kernel size and the number of filters are used by us are giving the best result. More than 360 node features more prone to overfit. We use Tanh activation function. We also experiment with different input length get best result with ${l_{1}}$ = 5 and ${l_{2}}$ = 10.  

We have shown our long-term prediction result for 560ms and 1000ms on the Human 3.6M dataset in table 6. Here, we have also outperformed almost all of the baseline methods except for discussion pose 560ms. We have used the same input length frame 10 and output length frame 25 to capture long-term prediction. Note on average, we get a better result for 560 and 1000ms.

\section{Conclusions and recommendations for future works}
Human Motion Prediction plays an important role in action recognition and autonomous driving. Earlier it is addressed as a sequence to sequence problem where RNNs and GRUs are used extensively. We have discussed problems with RNNs architecture. As we have shown adding a residual connection between inception block and input removes discontinuity between previous and next predicted frame and outperforms baseline models in both short and long term prediction. In our work, we use Inception Residual
Block(IRB) to detect temporal features and feed to Graph Convolution Network. GCN learns spatial features of the graph and predicts human motion. In future work,  multiple IRBs can be used and tested with other datasets.

\bibliographystyle{IEEEtran}
\bibliography{bib}

\end{document}